\documentclass[letterpaper]{article} 
\usepackage{aaai25}  
\usepackage{times}  
\usepackage{helvet}  
\usepackage{courier}  
\usepackage[hyphens]{url}  
\usepackage{graphicx} 
\usepackage{natbib}  
\usepackage{caption} 
\usepackage[linesnumbered,ruled,vlined]{algorithm2e}
\SetKwInput{KwInput}{Input}                
\SetKwInput{KwOutput}{Output}              

\usepackage{soul}
\usepackage{url}
\usepackage{mathtools}
\usepackage[utf8]{inputenc}
\usepackage[]{graphicx}
\usepackage{amsmath}
\usepackage{booktabs}
\usepackage{multirow}
\usepackage{algorithmic}
\usepackage[switch]{lineno}
\usepackage{xcolor}
\usepackage{amsfonts,amssymb}
\usepackage[bb=dsserif]{mathalpha}
\usepackage{bm}
\usepackage{anyfontsize}
\usepackage{caption}
\newcommand{\name}{{\tt CUQDS}}

\def\mathbi#1{\textbf{\em#1}}

\title{CUQDS: Conformal Uncertainty Quantification under Distribution Shift for Trajectory Prediction}
\author{
    Huiqun Huang\textsuperscript{\rm 1},
    Sihong He\textsuperscript{\rm 2}, Fei Miao\textsuperscript{\rm 1}\thanks{Corresponding author}\\
}
\affiliations{
    \textsuperscript{\rm 1}University of Connecticut\\
    \textsuperscript{\rm 2}University of Texas at Arlington\\
\{huiqun.huang, fei.miao\}@uconn.edu, sihong.he@uta.edu
}

\begin{document}

\maketitle

\begin{abstract}
Trajectory prediction models that can infer both future trajectories and their associated uncertainties of the target vehicles are crucial for safe and robust navigation and path planning of autonomous vehicles. However, the majority of existing trajectory prediction models have neither considered reducing the uncertainty as one objective during the training stage nor provided reliable uncertainty quantification during inference stage, especially under potential distribution shift. Therefore, in this paper, we propose the \textbf{C}onformal \textbf{U}ncertainty \textbf{Q}uantification under \textbf{D}istribution \textbf{S}hift framework, \name{}, to quantify the uncertainty of the predicted trajectories of existing trajectory prediction models under potential data distribution shift, while improving the prediction accuracy of the models and reducing the estimated uncertainty during the training stage. Specifically, \textbf{\name{}} includes 1) a learning-based Gaussian process regression module that models the output distribution of the base model (any existing trajectory prediction neural networks) and reduces the estimated uncertainty by an additional loss term, and 2) a statistical-based Conformal P control module to calibrate the estimated uncertainty from the Gaussian process regression module in an online setting under potential distribution shift between training and testing data.  Experimental results on various state-of-the-art methods using benchmark motion forecasting datasets demonstrate the effectiveness of our proposed design. 
\end{abstract}

\section{Introduction}\label{sec:introduction}
Accurately and efficiently predicting the future trajectories of the target vehicles is pivotal for ensuring the safety of path planning and navigation of autonomous systems in dynamic environments~\cite{hu2023planning,kedia2023integrated}. However, trajectory uncertainly due to the changing external surrounding environments (e.g., road networks, neighborhood vehicles, passengers, and so on) or intrinsic intention changes of drivers, can lead to both trajectory \textit{distribution shift} that change over time and \textit{overconfidence} in the output of trajectory prediction models. It remains challenging to predict the distribution of the future trajectories of the target vehicles (rather than focusing solely on point estimates), and to quantify the uncertainty of the output trajectories under potential distribution shift between training and testing data.

Various uncertainty quantification methods have been proposed for trajectory prediction. One popular approach is to estimate the distribution (e.g., Laplace distribution or Gaussian distribution) of future trajectories of target vehicles by direct modeling methods~\cite{zhou2022hivt,mao2023leapfrog,zhu2023ipcc,salzmann2020trajectron++,chen2023unsupervised,tang2021collaborative}. However, this method often overlooks the impact of \textit{model limitation} (the restricted ability of models to represent the real-time trajectories data) and the consequence overconfident uncertainty estimation in the inference stage. 
Another popular approach is to calibrate the preliminary estimated uncertainty in the inference stage by the statistical-based methods. Specifically, methods such as split conformal prediction (CP)~\cite{shafer2008tutorial,lindemann2023safe,lekeufack2023conformal} provide confidence intervals that guarantee to contain the ground truth target data with a predefined probability. However, they are not applicable to the situations when there is distribution shift between the training and testing data.
\begin{figure}[ht]
 \centering
 \includegraphics[width=0.7\linewidth]{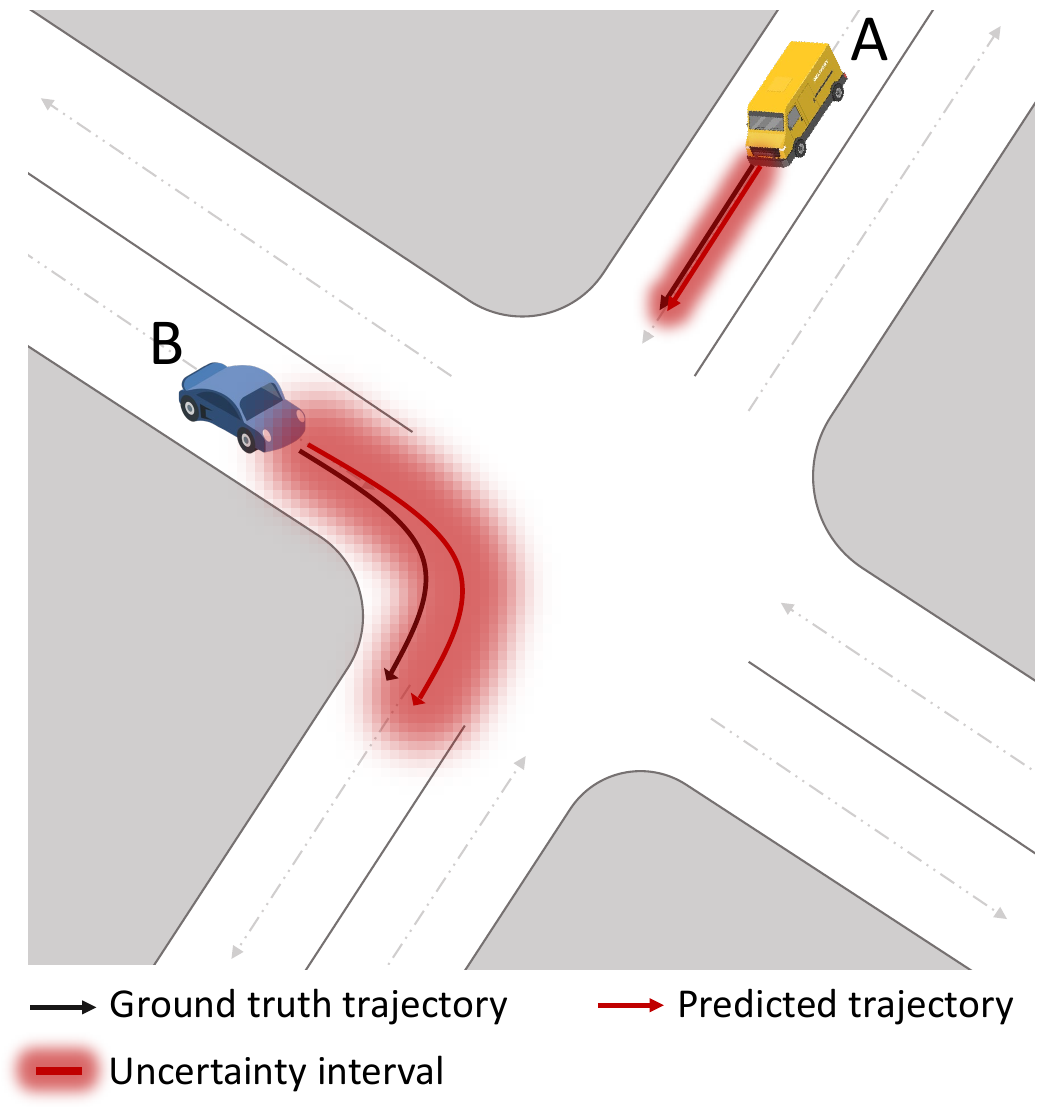} 
  \caption{Illustration of the importance of a good uncertainty interval estimation for the predicted trajectories of the target vehicles.}
  \vspace{-8pt}
  \label{fig:uncertainty_interval}
\end{figure}


In this paper, we propose \textbf{\name{}}, i.e., \textbf{C}onformal \textbf{U}ncertainty \textbf{Q}uantification under \textbf{D}istribution \textbf{S}hift framework, to quantify the output uncertainty of existing trajectory prediction models under distribution shift, while improving the prediction accuracy of the models and reducing the estimated uncertainty during the training stage. The proposed \textbf{\name{}} framework integrates a learning-based Gaussian process regression~\cite{rasmussen2006gaussian} module with a statistics-based conformal P control module~\cite{angelopoulos2023conformal}. Specifically, 
in the training stage, \textbf{\name{}} adopts the Gaussian process regression module to estimate the output distribution of base model. The variance of the output distribution quantifies the uncertainty of the predicted trajectories. We further use the output distribution to construct uncertainty interval that guarantees to cover the true target trajectory with a predefined probability in the long-run. As shown in Fig.~\ref{fig:uncertainty_interval}, a good uncertainty interval should tend to be narrow when the predicted trajectory closely aligns with the ground truth trajectory (e.g., vehicle A), whereas it should be widen when the output trajectory deviates (e.g., vehicle B). To achieve this, we introduce additional loss term to reduce the estimated uncertainty in the training stage, ensuring narrow uncertainty interval while covering the true target trajectory with a predefined probability. 
In the inference stage, \textbf{\name{}} calibrates the estimated output uncertainty upon potential distribution shift by a statistics-based conformal P control module. Different from standard conformal prediction method and its variants who use fixed conformal quantile in the inference stage and violate the distribution shift assumption, we first initialize the conformal quantile using the validation data and update it after every prediction step during the inference stage. 

The key contributions of this work are summarized as follows:

\begin{enumerate}
  \item We propose \textbf{\name{}} framework that integrates both the learning-based and statistical-based modules to provide uncertainty quantification for trajectory prediction under distribution shift. The proposed learning-based module is trained alongside the base model to quantify its output uncertainty. In the training stage, the main objectives are to enhance prediction accuracy of base model and reduce the estimated uncertainty by incorporating an additional loss term.

  \item We introduce the statistical-based conformal P control module to calibrate the estimated output uncertainty from the learning-based module under potential distribution shift during the inference stage. To alleviate the impacts of data distribution shift on uncertainty estimation, the conformal quantile is first initialized using the validation data and keep updating after each prediction step in the inference stage.
  

  \item We validate the effectiveness of our proposed framework on the Argoverse 1 motion forecasting dataset~\cite{chang2019argoverse} and five state-of-the-art baselines. Compared to base models without our \textbf{\name{}}, the experiment results show that our approach improves prediction accuracy by an average of 7.07\% and reduce uncertainty of predicted trajectory by an average of 25.41\%.  

\end{enumerate}


\section{Related Work}\label{sec:related_work}

\textbf{Modeling the distribution of the trajectory} of the target vehicles, instead of solely focusing on the point estimation of the future trajectory, proves to be an efficient approach~\cite{chai2019multipath,deo2018convolutional,phan2020covernet,zeng2021lanercnn} to avoid missing potential behavior in trajectory prediction methods. Early works sample multiple potential future trajectories~\cite{liang2020learning,ngiam2021scene,gupta2018social,rhinehart2018r2p2,rhinehart2019precog,tang2019multiple} to approximate the predicted trajectory distribution. 
However, these approaches still rely on limited point estimation and suffer from overconfidence upon trajectory prediction in real-world settings. To mitigate these issues, recent works directly estimate the distribution of future trajectories by fitting the (mixture) distribution models based on the embedded features of input trajectories~\cite{zhou2022hivt,varadarajan2022multipath++,cui2019multimodal,shi2023mtr++}. 
However, these works mostly train the (mixture) distribution models by the MLP-based predictor in the last layer of the model. Such designs fail to estimate an accurate enough distribution of future trajectories under data distribution shift.

\textbf{Uncertainty quantification in trajectory prediction} 
is challenging and usually solved by 
two categories of methods: direct modeling and statistical-based methods. Specifically, direct modeling assumes the target data follows specific distribution, designs the learning-based neural networks for distribution estimation, and introduces the corresponding loss function to model the uncertainty directly. To achieve this, existing works usually assume the data follows Gaussian distribution~\cite{mao2023leapfrog} or Laplace distribution~\cite{gu2024producing} and model the corresponding distribution by learning-based designs. However, estimating rigorous uncertainty by direct modeling can be challenging, as the model may easily overfit the training dataset such that invalid in the inference stage under data distribution shift. \textbf{Conformal prediction}~\cite{shafer2008tutorial,angelopoulos2023conformal} stands out as a method in statistical-based inference that has proven effective in constructing predictive sets, ensuring a certain probability of covering the true target values~\cite{ivanovic2022propagating,xu2014motion,sun2024conformal,xu2023model,stankeviciute2021conformal}. However, the assumption of identical distribution across all datasets invalidates the standard conformal prediction under distribution shift. To address this, \textbf{adaptive conformal prediction} methods~\cite{gibbs2021adaptive,stanton2023bayesian} have been proposed to achieve the desired coverage frequency under distribution shift by leveraging the distance between new queries and the training dataset. However, these approaches either lack base model uncertainty estimation or fail to incorporate mechanisms for reducing uncertainty during the training process, as achieved by our proposed \name{}.


\begin{figure*}
 \centering
 \includegraphics[width=0.9\linewidth]{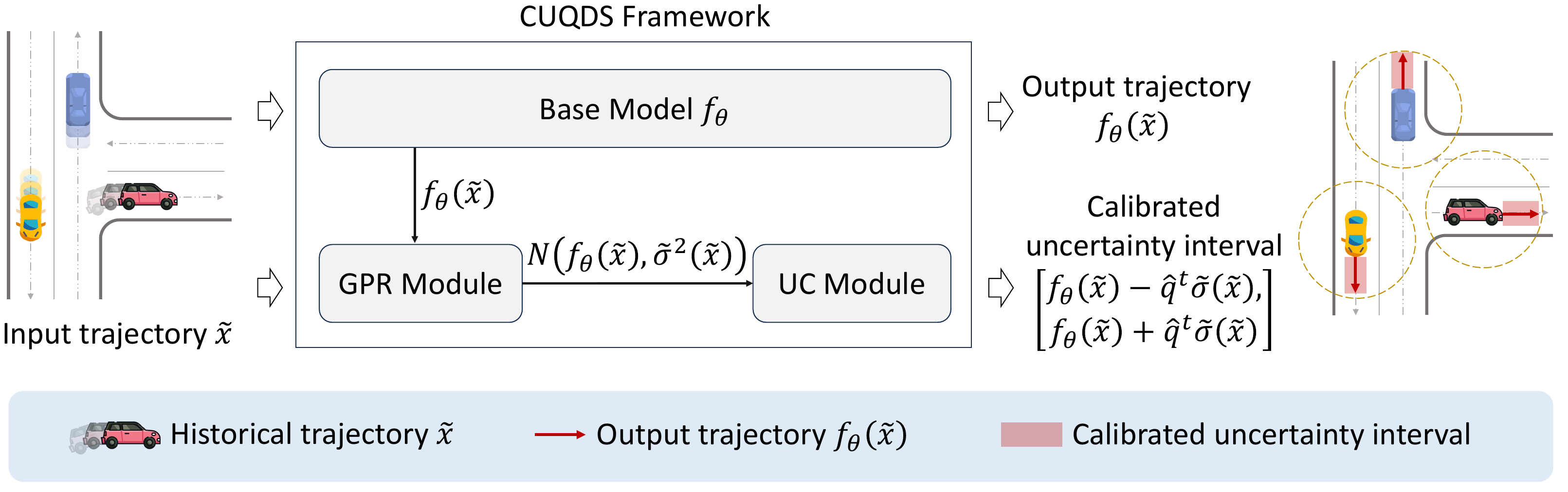} 
  \caption{Our \textbf{\name{}} models the conditional output distribution $\tilde Y|\tilde x, \mathcal{D} \sim \mathcal{N}\left(f_{\theta}\left(\tilde x\right), \tilde{{\sigma}}^{2}\left(\tilde x\right)\right)$ of base model $f_{\theta}$ by the Gaussian process regression (GPR) module, and provides the correspond calibrated uncertainty interval $\left[f_{\theta}\left(\tilde x\right)-\hat{q}^t\tilde{{\sigma}}\left(\tilde x\right), f_{\theta}\left(\tilde x\right)+\hat{q}^t\tilde{{\sigma}}\left(\tilde x\right)\right]$ for the predicted trajectories by the uncertainty calibration (UC) module.}
  \label{fig:framework}
\end{figure*}

In this study, we propose the \name{} framework to provide output distribution for the predicted trajectories of base model under distribution shift, while improving the prediction accuracy of base model and reducing the estimated uncertainty by introducing additional loss term. In particular, \textbf{\name{}} adopts a Gaussian process regression module to estimate the output uncertainty of base model, then utilizes a statistical-based conformal P control module to calibrate this output uncertainty by taking into account the model performance on recent trajectory data. Moreover, different from standard conformal prediction methods who use the fixed conformal quantile in the inference stage, we first initialize the conformal quantile using the validation data and update it after every prediction in the inference stage to alleviate the data distribution shift problem.




\section{Methodology}\label{sec:methodology}
\subsection{Problem Formulation} \label{subsec:problem_formulation}
Suppose we have a training dataset $\mathcal{D}_1 = \{(x_i, y_i)\}_{i=1}^{N_1}=\{(x^t, y^t)\}_{t=L-1}^{T_1}$, a validation dataset $\mathcal{D}_2 = \{(x_i, y_i)\}_{i=1}^{N_2}=\{(x^t, y^t)\}_{t=L-1}^{T_2}$, and a testing dataset $\mathcal{\tilde D} = \{(\tilde x_i, \tilde y_i)\}_{i=1}^{N_3}=\{(\tilde x^t, \tilde y^t)\}_{t=L-1}^{T_3}$. $N_1$, $N_2$, and $N_3$ are the number of data samples, and $T_1$, $T_2$, and $T_3$ are the time periods of each dataset. $T_1 \bigcap T_2 \bigcap T_3 = \varnothing$. In this paper, we assume that training/validation datasets are drawn from the same distribution, while the distribution of testing dataset shifts over time. As shown above, we use two ways to represent each dataset, where $(x_i, y_i)$ (or $(\tilde x_i, \tilde y_i)$) denotes the random input-output trajectory pair within each dataset indexed by $i$, and $(x^t, y^t)$ (or $(\tilde x^t, \tilde y^t)$) is time series data and represents the input-output trajectory pair of the current time step $t$. Both $x_i$ and $x^t$ denote the input historical trajectory during the past $L$ historical time steps and are sampled from domain $\mathcal{X} \in \mathbb{R}^{L \times D}$, and $y_i$ and $y^t$ denote the corresponding target trajectory during the $J$ following time steps from domain $\mathcal{Y} \in \mathbb{R}^{J \times D}$. $D$ is the dimension of target features. We will omit the index $i$ or $t$ when there is no conflict.

We assume that we have a trajectory prediction model $f$ with parameters $\theta$. We call this model $f_{\theta}$ a base model and it can be implemented as different structures of deep neural network~\cite{zhou2022hivt,liu2021multimodal,zhou2023query,liang2020learning,zhong2022aware}. 
The model $f_{\theta}$ is trained using the dataset $\mathcal{D}$, where the target trajectory, denoted as $Y$, is conditioned on input $x$ and the dataset $\mathcal{D}$. We model the conditional distribution of $Y$ given $x$ and $\mathcal{D}$ as a Gaussian distribution, where $Y | x, \mathcal{D} \sim \mathcal{N}\left(\mu(x), \sigma^2(x)\right)$. In this formulation, $\mu(x)$ and $\sigma^2(x)$ are functions that map the input $x$ to the mean and variance of the Gaussian distribution, respectively. We treat the predicted trajectory of the base model $f_{\theta}$ as the mean $\mu(x)$ directly. Our main task in this study is to estimate the variance $\sigma^2(x)$ to denote the output uncertainty.

We propose the Conformal Uncertainty Quantification framework, \textbf{\name{}}, as illustrated in Fig.~\ref{fig:framework}, to quantify the uncertainty of the predicted trajectory of base model under potential distribution shift. The main objectives of this framework are to improve prediction accuracy of base model and reduce the output uncertainty. The framework introduces two modules,
\textit{1)} In particular, in the training and validation stages ( Alg.~\ref{alg:training_and_validation_stages}), we propose a learning-based Gaussian process regression module to approximate the conditional output distribution $(Y|x, \mathcal{D})$ of base model by estimating $\left(Y|x, \theta \sim \mathcal{N}\left(\hat{\mu}\left(x\right), \hat{{\sigma}}^{2}\left(x\right)\right)\right)$ based on $f_\theta$. 
\textit{2)} In the testing or inference stage (Algorithm.~\ref{alg:testing_stage}), we propose a statistical-based conformal P control module to calibrate the output distribution $\tilde Y|\tilde x, \theta \sim \mathcal{N}\left(\tilde{\mu}\left(\tilde x\right), \tilde{{\sigma}}^{2}\left(\tilde x\right)\right)$ by considering the potential distribution shift between the training and testing datasets and the performance limitation of $f_{\theta}$ on current time series trajectory input. The conformal quantile of the module is first initialized using the validation data and keep updating after each prediction during the inference stage. We build the calibrated uncertainty interval and ensure it covers the true target trajectory with a predefined probability in long-run. 

To summarize,  during training and validation stage, our goal is to find the parameters $[\theta, \omega]$ such that minimizing the loss function $\mathcal{L}$ on training data: 
\begin{equation}\label{eq:optimization}
\centering
  \begin{aligned}
  \left[{\theta}, {\omega}\right]= \underset{\theta, {\omega}}{\texttt{arg} \, \texttt{min}} \; \mathcal{L}\left(\theta, \omega | \mathcal{D} \right).
  \end{aligned}
\end{equation}
The loss $\mathcal{L}$ is a weighted combination of base model's loss $\mathcal{L}_1(\theta)$ and Gaussian processing regression model's loss $\mathcal{L}_2(\omega, \theta)$. 
\begin{equation}\label{eq:loss_function}
\centering
  \begin{aligned}
   \mathcal{L}=w_1\mathcal{L}_1(\theta) +w_2\mathcal{L}_2(\omega, \theta), 
  \end{aligned}
\end{equation}
where $w_1 \in \mathbb{R}$ and $w_2 \in \mathbb{R}$ are the weights adjusting the influence of two loss terms, respectively. During inference stage, we will calibrate the estimated uncertainty of the predicted trajectory.

\subsection{Preliminary}\label{subsec:preliminary}
$\bullet$~\textbf{Gaussian Process Regression} often acts as the surrogate model to estimate the output distribution of existing models~\cite{erlygin2023uncertainty}. It introduces additional loss term to guild the learning process of existing models and reduce the output uncertainty. Gaussian process regression assumes any combinations of data samples follows different join distribution and thus good at capturing the nonlinear relationship among time series data samples. 

Given an existing deep neural network model $f_{\theta}$, Gaussian process regression views this model as a black box and aims to estimate the output distribution $(Y|x, \theta \sim \mathcal{N}\left(\hat{\mu}\left(x\right), \hat{{\sigma}}^{2}\left(x\right)\right))$ corresponding to the inputs $x$. We treat the output of model $f_{\theta}\left(\cdot\right)$ on $x$ as the mean value $\hat{\mu}\left(x\right)$ of the estimated output distribution and the correspond variance $\hat{{\sigma}}^{2}\left(x\right)$ as the uncertainty of $f_{\theta}\left(x\right)$. Gaussian process regression also assumes that the observed values $y$ differ from the function values $f_{\theta}(x)$ by additive noise $\epsilon$, and further assumes that this noise follows an independent, identically distributed Gaussian distribution with zero mean and variance of ${\sigma}_{\epsilon}^{2}$. Now the prediction process is formulated as $Y|_x = \hat \mu(x) + \epsilon$, where $(x, y)$ is any data samples from $\mathcal{D}$. Then we have:
\begin{align}
\label{eq:training_predictive_dis}
   &Y|_x \sim N\left(\hat{\mu}\left(x\right), \frac{{\sum}_{i=1}^{N_1} k\left(x, x_i\right)}{N_1}+{\sigma}_{\epsilon}^{2}I\right),
\end{align}
where $k(\cdot, \cdot)$ is the user-defined kernel function or the covariance function. For example,  $k\left(x, x'\right)$ quantifies the similarity between trajectory input data $x$ and $x'$. 

$\bullet$~\textbf{Conformal Prediction}~\cite{shafer2008tutorial,angelopoulos2023conformal} is a statistical-based method to construct predictive sets for any model, ensuring a certain probability of covering the true target values. It assumes identical distribution across all datasets. Given the base model $f_{\theta}$ whose output distribution follows $(Y|x, \theta \sim \mathcal{N}\left(\hat{\mu}\left(x\right), \hat{{\sigma}}^{2}\left(x\right)\right))$, conformal prediction converts the heuristic notion of uncertainty (e.g. the estimated standard deviation $\hat{{\sigma}}\left(x\right)$) into rigorous prediction intervals of the form $\left(\hat{\mu}\left(x\right)\pm q\hat{{\sigma}}\left(x\right)\right)$, where $q$ is the estimated conformal quantile from conformal prediction using validation dataset.

Given the validation dataset $\mathcal{D}_2$ which has $N_2$ data samples, conformal prediction follows the following steps: \textit{1)} Define a conformal score function $s(x,y) \in \mathbb{R}$ (smaller scores encode better agreement between x and y). \textit{2)} Compute the conformal quantile $q$ as the $\frac{\lceil (N_2+1)(1-\alpha) \rceil}{N_2}$ quantile of the validation scores $S=\{s(x_i,y_i)\}_{i=1}^{N_2}$. $\alpha \in [0,1]$ is the user-defined error rate. \textit{3)} Then for every unseen data sample $(\tilde x, \tilde y)$, conformal prediction forms the prediction set $\mathcal{C}(\tilde 
y|s(\tilde x, \tilde y) \leq q)$. Conformal prediction guarantees that the prediction set contains the true target value with a probability at least $1-\alpha$.

\subsection{Gaussian Process Regression Module} \label{subsec:output_distribution_modeling}
Existing literature of trajectory prediction models~\cite{zhou2022hivt,zhou2023query,liang2020learning,zhong2022aware} focuses on providing point estimates of future trajectories. However, the uncertainty of the future trajectories due to the changing environment or the intrinsic intention changes of drivers can lead to significant distribution shift and overconfident trajectory prediction. Such distribution shift and overconfident prediction can greatly impact the subsequent decision-making processes, such as robust path planning~\cite{hu2023planning,kedia2023integrated}. Hence, we propose to consider output uncertainties from both the modeling limitation of the model $f_\theta$ upon current trajectory inputs and the noise inherent in the trajectory data. 

\begin{algorithm}[tb]
\caption{Training \& validation stage}
\label{alg:training_and_validation_stages}
\textbf{Input}: a base model $f(\cdot)$ with an initialized parameter $\theta$ and a covariance function $k(\cdot)$ with an initialized parameter $\omega$, error rate $\alpha \in [0, 1]$, score function $s(\cdot)$, total number of training epochs $epo$.\\
\textbf{Data}: training dataset $D_1=\{x_i, y_i\}_{i=1}^{N_1}$, validation dataset $D_2=\{x^t, y^t\}_{t=1}^{T_2}$.\\
\textbf{Output}: well-trained base model $f_\theta(\cdot)$, updated statistic $\hat{q}^1$, score set $S={\{s\left(x^t, y^t\right)\}}_{t=L-1}^{T_2}$ and error set $E={\{e^t\}}_{t=L-1}^{T_2}$ estimated from the validation data.\\
 Initialization: $\hat{q}^0=1$.\\
  \For{epoch in $epo$}    
    {   
        Training \name{}$_{\theta, \omega}$ with loss $\mathcal{L}$ and training dataset $D_1$.\\
        Update $\{\theta, \omega\} \leftarrow  \underset{\theta, {\omega}}{\texttt{arg} \, \texttt{min}} \; \mathcal{L}\left(\theta, \omega | \mathcal{D} \right)$ \\
        Initialize score set $S=\{\}$,  error set $E=\{\}$.
        \\
        
        \For{$t=1:T_2$}
        {   
            $f_{\theta}(x^t), \hat\sigma(x^t) \leftarrow \name{}_{\theta, \omega}(x^t)$\\
            Compute score $s^t = s(x^t, y^t)$ by Eq.~\ref{eq:conformal_score}.\\
            Compute conformal prediction set $\mathcal{C}^t$ by Eq.~\ref{eq:uncertainty_interval_testing}.\\ 
            Compute $e^t = \mathbbb{1}_{\tilde y^t \notin \mathcal{C}^t}$.\\
            $S \leftarrow s^t$, $E \leftarrow e^t$.\\
           $\eta = \beta \max\left(S \right)$, then $\hat{q}^t = \hat{q}^{t-1} + \eta (\bar{E} - \alpha)$.
        }
    }
Update $\hat{q}^1 \leftarrow \hat{q}^{T_2}$.
\end{algorithm}

In this study, we introduce to utilize the Gaussian process regression method as introduced in preliminary to estimate the preliminary output distribution of existing trajectory prediction models. In our case, the base model $f_{\theta}$ can be any existing deep neural network structure in the literature, examples include~\cite{zhou2022hivt,liu2021multimodal,zhou2023query,liang2020learning,zhong2022aware}). Moreover, in real-world settings of autonomous vehicles, the collected historical trajectories of the target vehicles inevitably contain noisy data during perception and object tracking steps. Such noise usually hardly captured by the trajectory prediction models such that greatly impacts the output trajectory and the corresponding uncertainty. To mitigate this problem, we model such noisy impact as $\epsilon \in \mathbb{R}^{D}$ as in the standard Gaussian process regression method. 

\subsubsection{Learnable Kernel Function}
In this paper, we define the covariance function as the radial basis function  
\begin{equation}\label{eq:covariance_function}
\centering
  \begin{aligned}
   k\left(x, x'\right)={l}_{1}^{2}exp\left(-\frac{{\left(x-x'\right)}^2}{2{l}_{2}^{2}}\right),
  \end{aligned}
\end{equation}
where \(l_1 \in \mathbb{R}\) and \(l_2 \in \mathbb{R}\) are the parameters to be trained. The closer \(x\) and \(x'\) are, the higher the value of \(k(x, x')\), reaching its maximum value of \(l_1^2\). 

\subsubsection{Eigenvector Inducing Variables}
As shown in Eq.~\ref{eq:training_predictive_dis}, the covariance $\hat{{\sigma}}^{2}(x)$ of standard Gaussian process regression method is typically optimized based on the whole training data.
However, this can be computationally expensive when the number of training samples (\(N_1\)) is large, given that the entire Gaussian process regression model requires \(O(N_1^3)\) computational complexity and \(O(N_1^2)\) memory complexity. To alleviate such complexity while maintaining the effectiveness of the Gaussian process regression module, we propose a modification to approximate the covariance $\hat{{\sigma}}^{2}(x)$ in Eq.~\ref{eq:training_predictive_dis} by 
extracting $M \in \mathbb{R}$, $M \ll N_1$, inducing variables $\{v_i\}_{i=1}^{M}$, $v_i \in \mathbb{R}^{L \times D}$ by Principal Component Analysis (PCA). These $M$ inducing variables summarize the key information of training dataset. More specifically, we first standardize each input trajectory data sample $x_i \in \mathbb{R}^{L \times D}$ in training dataset $\mathcal{D}_1$ by its own mean and standardization along time dimension. Then we build the $N_1 \times N_1$ covariance matrix by $cov(x, x') = \mathbi{E}[(x-\mathbi{E}(x))(x'-\mathbi{E}(x'))]$. The further steps follow the standard PCA processes.

\subsubsection{Inference under Distribution Shift}
Through the Gaussian process regression with a learnable kernel function, we are able to estimate the target trajectory's distribution that in the training/validation dataset. However, under potential distribution shift between training and testing data in real-world settings, solely keep using the information learn from the training/validation datasets will lead to high generalization error. To take into account the potential distribution shift between the dataset $\mathcal{D}=\{\mathcal{D}_1, \mathcal{D}_2\}$ and the testing dataset $\mathcal{\tilde{D}}$, we instead estimating the conditional output distribution of base model in the testing stage by $\Tilde{Y}|\tilde{x},\mathcal{D}  \sim \mathcal{N}(\tilde\mu(\tilde{x}), \tilde\sigma(\tilde{x}))$, where $(\tilde{x}, \tilde{y}) \in \tilde{\mathcal{D}}$. 
Then we compute the variance of the conditional output distribution of base model in the testing stage by
\begin{align}
{\tilde\sigma}^{2}(\tilde{x})&=k(\tilde{x},\tilde{x}) - K_{\tilde{x}M} [K_{MM} + \sigma^2_\epsilon I]^{-1} (K_{\tilde{x}M})^\intercal, 
\end{align}
where 
$K_{\tilde{x}M}=\left(k\left(\tilde{x}, v_i\right),\dots, k\left(\tilde{x}, v_M\right)\right)^\intercal$ and $[K_{MM}]_{ij}=k\left(v_i, v_j\right)$. We then construct the uncertainty interval as $[f_{\theta}(\tilde{x})-\tilde\sigma(\tilde{x}), f_{\theta}(\tilde{x})+\tilde\sigma(\tilde{x})]$, indicating there is a high probability that it will cover the true target trajectories.

To find the optimal parameters for the covariance function and the noise $\epsilon$, and reduce the estimated uncertainty, we introduce a new loss term
$\mathcal{L}_2$:
\begin{align}
    \label{eq:loss^term_l2}
    \mathcal{L}_2(\omega, \theta) = \frac{1}{N_1} {\sum}_{i=1}^{N_1} ( -\frac{1}{2}e_i^{\top}{\left[\bar{K}_{x_{i}M}+{\sigma}^{2}_{\epsilon}I\right]}^{-1}e_i\\-
    \frac{1}{2}log\left|\bar{K}_{x_{i}M}+{\sigma}^{2}_{\epsilon}I\right|) + \frac{1}{2 N_1} log 2\pi ),
\end{align}

where $\omega = [l_1, l_2, {\sigma}_{\epsilon}]$ contains all trainable parameters in the Gaussian process regression module, $\bar{K}_{x_{i}M}=\frac{1}{M}{\sum}_{j=1}^{M}k(x_i, v_j)$, and $e_i = y_i - f_{\theta}\left(x_i\right)$.

\subsection{Uncertainty Quantification under Distribution Shift through Calibration} \label{subsec:cp_under_drift}
The learning-based Gaussian process regression module is prone to overfit the training data and provide overconfident uncertainty estimation under distribution shift. 
To solve this, we propose a statistical-based conformal P control module to calibrate the output uncertainty from the Gaussian process regression module by considering the performance of base model under distribution shift. The idea of this method is based on both the conformal prediction~\cite{shafer2008tutorial} and the P control in~\cite{angelopoulos2023conformal}. However, different from standard conformal prediction, we assume the training/validation data and testing data follow different distributions, and the distribution of testing data shifts over time. Instead of using the fixed conformal quantile $q$ as in standard conformal prediction, our proposed module updates the conformal quantile $q^t$ in time step $t$ based on model performance under distribution shift.

In the inference stage, we aim to achieve a long-run average coverage rate in time, ensuring that the calibrated uncertainty interval covers the true target trajectories with an average probability of $1 - \alpha$, namely $\sum_{t=1}^{T}\texttt{err}^t/T \rightarrow \alpha$ as $T \rightarrow \infty$. $\texttt{err}^t = \mathbbb{1}(y^t \notin \mathcal C^t)$ and $\alpha \in (0,1)$ is a predefined target error rate threshold. 
 


\subsubsection{Uncertainty Calibration in \name{} Framework}
To update the conformal quantile $q^t$ under distribution shift, we first initialize the conformal quantile using the validation data as shown in Alg.~\ref{alg:training_and_validation_stages}. 
For each data sample $(x^t, y^t)$ in validation data \(D_2\), we first compute the corresponding conformal score \( s(x^t, y^t) \) and the conformal prediction set \( \mathcal{C}^t \) as standard conformal prediction. We define the score function $s(x,y)$ to evaluate the model performance for each data sample 
\begin{equation}\label{eq:conformal_score}
\centering
  \begin{aligned}
   s(x,y) = \frac{|y - f_{\theta}(x)|}{\sigma(x)},
  \end{aligned}
\end{equation}
where $y$ is the ground truth, $f_{\theta}(x)$ and $\sigma(x)$ are respectively the mean and std of the output distribution of the base model.

Moreover, we introduce an additional error set \( E \) to record the coverage errors of uncertainty intervals. The error \( e^t \) is set to 1 if \( y^t \) is outside of \( \mathcal{C}^t \), indicating a prediction error, and 0 otherwise. The estimated conformal score \( s(x^t, y^t) \) is added to the score set \( S \), and the error \( e^t \) is added to the error set \( E \). The conformal quantile \(\hat{q}^t\) is then updated based on both the existing conformal scores and coverage errors. The updating rule involves a learning rate \( \eta = \beta \max(S) \) and adjusts \(\hat{q}^t\) according to the formula 
$$\hat{q}^t = \hat{q}^{t-1} + \eta (\bar{E} - \alpha),$$
where \( \bar{E} \) is the average coverage error in the coverage error set and \( \alpha \) is the predefined error rate. This step allows the model to adapt its uncertainty estimation based on the observed performance during validation. 

During the inference stage, the conformal P control module inherits the conformal quantile, the score set \({S}\), and the error set \({E}\) from the final validation iteration. The conformal quantile updating steps in the inference stage as shown in Alg.~\ref{alg:testing_stage} are similar as in the validation stage. For each time step \(t\) in the inference stage, we calibrate the output uncertainty and construct the uncertainty interval by using the updated conformal quantile \(\hat{q}^t\),
\begin{equation}\label{eq:uncertainty_interval_testing}
\centering
  \begin{aligned}
   \mathcal{C}^t = \left[f_{\theta}( \tilde x^t)-\hat{q}^t {\tilde\sigma}(\tilde{x}^t), f_{\theta}( \tilde x^t)+\hat{q}^t {\tilde\sigma}(\tilde{x}^t)\right].
  \end{aligned}
\end{equation}

\begin{algorithm}[tb]
\caption{Testing stage}
\label{alg:testing_stage}
\textbf{Input}: the well-trained framework $\name{}_{\theta, \omega}$, the conformal quantile $\hat{q}^1$, the error set $E$, the conformal score set ${S}$, the error rate $\alpha$ from the training stage, score function $s(\cdot)$.\\
\textbf{Data}: testing data $\tilde D=\{\left(\tilde x^t, \tilde y^t\right)\}_{t=L-1}^{T_3}$.\\
\textbf{Output}: predicted trajectory and the correspond calibrated uncertainty interval.\\
  \For{$t=1:T_3$}
        {   
            $f_{\theta}(\tilde x^t), \hat\sigma(\tilde x^t) \leftarrow \name{}_{\theta, \omega
            }(\tilde x^t)$\\
            Calibrate std $\tilde\sigma( \tilde x^t)$ by $\hat{q}^{t-1} \tilde\sigma( \tilde x^t)$.\\
            Compute score $s(\tilde x^t, \tilde y^t)$ by Eq.~\ref{eq:conformal_score}.\\
            Compute conformal prediction set $\mathcal{C}^t$ by Eq.~\ref{eq:uncertainty_interval_testing}.\\ 
            Compute $e^t = \mathbbb{1}_{\tilde y^t \notin \mathcal{C}^t}$.\\
            $S \leftarrow s^t$, $E \leftarrow e^t$.\\
           $\eta = \beta \max\left(S \right)$, then $\hat{q}^t = \hat{q}^{t-1} + \eta (\bar{E} - \alpha)$.
        }
\end{algorithm}

\section{Experiment}\label{sec:experiment}

\subsection{Experimental Setups} \label{subsec:experimental_setups}

\textbf{Dataset \& Key Setups:} We use the Argoverse 1 motion forecasting dataset~\cite{chang2019argoverse} to verify the efficacy of our approach. This dataset collects trajectory data from Miami and Pittsburgh, with a sample rate of 10 Hz.
Given that the ground truth future trajectories are not provided in this official test sequences but are essential to our \textbf{\name{}} in the testing stage to update the conformal quantile $\hat{q}^t$, we repartition the sequences. In particular, we split the 205,942 official training sequences into 166,470 training data and 39,472 validation data, and use the official 39,472 validation sequences as testing data in this study.

In all experiments of this study, $\alpha$ is set as 0.1. We implement existing base models by using their default settings, unless otherwise specified. The host machine is a server with IntelCore i9-10900X processors and four NVIDIA Quadro RTX 6000 GPUs.

\textbf{Prediction Accuracy Evaluation Metrics:} Following the standard evaluation protocol, we utilize metrics including minimum Average Displacement Error ($\texttt{minADE}_k$), minimum Final Displacement Error ($\texttt{minFDE}_k$), and Miss Rate ($\texttt{MR}_k$) to evaluate the prediction accuracy of the model. 
As common, $k$ is selected as 1 and 6.

\textbf{Uncertainty Evaluation Metrics:} To further verify the efficacy of our \textbf{\name{}} in reducing the predicted uncertainty to provide narrow while accurate uncertainty interval, we adopt the 
Negative Log-Likelihood (NLL)~\cite{feng2021review} to assess the level of uncertainty in the predicted distribution. Lower values connote a higher degree of precision in uncertainty estimation and narrower uncertainty interval.

\begin{table*}[]
\small
\centering
\renewcommand{\arraystretch}{1.2}
\resizebox{0.6\linewidth}{!}{\begin{tabular}{|l||ccc|ccc|}
\hline
Scheme  &  ${\texttt{minADE}}_6 \downarrow$  & ${\texttt{minFDE}}_6 \downarrow$  & ${\texttt{MR}}_6 \downarrow$   & ${\texttt{minADE}}_1 \downarrow$  & ${\texttt{minFDE}}_1 \downarrow$  & ${\texttt{MR}}_1 \downarrow$    \\ \hline
\texttt{HiVT} & 0.692 & 1.043 & 0.106 & 1.291 & 2.895 & 0.499  \\ \hline
\texttt{HiVT+SPCI} & 0.694 & 1.047 & 0.107 & 1.302 & 2.921 & 0.501  \\ \hline
\texttt{HiVT}+\textbf{\name} & \textbf{0.682} & \textbf{1.034} & \textbf{0.097} & \textbf{1.218} & \textbf{2.696} & \textbf{0.455} \\ \hline \hline

\texttt{LaneGCN} & 0.719 & 1.094 & 0.104 & 1.375 & 3.023 & 0.505 \\ \hline
\texttt{LaneGCN+SPCI} & 0.739 & 1.169 & 0.114 & 1.585 & 3.573 & 0.568 \\ \hline
\texttt{LaneGCN}+\textbf{\name} & \textbf{0.704} & \textbf{1.033} & \textbf{0.097} & \textbf{1.255} & \textbf{2.744} & \textbf{0.482}  \\ \hline \hline

\texttt{LBA} & 0.717 & 1.094 & 0.103 & 1.395 & 3.100 & 0.503  \\ \hline
\texttt{LBA+SPCI} & 0.719 & 1.098 & 0.104 & 1.402 & 3.100 & 0.523 \\ \hline
\texttt{LBA}+\textbf{\name} & \textbf{0.705} & \textbf{1.044} & \textbf{0.092} & \textbf{1.251} & \textbf{2.801} & \textbf{0.483} \\ \hline \hline

\texttt{LBF} & 0.720 & 1.098 & 0.104 & 1.530 & 3.467 & 0.549  \\ \hline
\texttt{LBF+SPCI} & 0.721 & 1.099 & 0.105 & 1.533 & 3.469 & 0.550  \\ \hline
\texttt{LBF}+\textbf{\name} & \textbf{0.714} & \textbf{1.091} & \textbf{0.097} & \textbf{1.455} & \textbf{3.188} & \textbf{0.522}\\ \hline\hline

\texttt{Trajectron++} & 0.764 & 1.121 & 0.123 & 1.602 & 3.713 & 0.589  \\ \hline
\texttt{Trajectron+++SPCI} & 0.732 & 1.079 & 0.119 & 1.593 & 3.572 & 0.571  \\ \hline
\texttt{Trajectron++}+\textbf{\name} & \textbf{0.693} & \textbf{1.101} & \textbf{0.112} & \textbf{1.491} & \textbf{3.404} & \textbf{0.548} \\ \hline

\end{tabular}}
\caption{Prediction results and performance comparison on testing dataset when with and without our \textbf{\name}.
}
\label{tab:argoverse_v1^testing_results}
\vspace{-0.1in}
\end{table*}

\begin{table}[]
\small
\centering
\renewcommand{\arraystretch}{1.2}
\vspace{-0.1in}
\resizebox{\linewidth}{!}{\begin{tabular}{|l||c|c||c|c|c|}
\hline
Scheme  &  \texttt{+SPCI (NLL)}  & \texttt{+\textbf{\name} (NLL)} &  \texttt{+SPCI (CR)}  & \texttt{+\textbf{\name} (CR)}   \\ \hline
\texttt{HIVT} & 21.354 & 16.354 & 0.705 & 0.832 \\ \hline
\texttt{LaneGCN} & 25.532 & 18.896 & 0.603 & 0.746 \\ \hline
\texttt{LBA} & 23.634 & 17.453 & 0.602 & 0.721 \\ \hline 
\texttt{LBF} & 23.723 & 17.535 & 0.614 & 0.716 \\ \hline 
\end{tabular}}
\caption{Compare the NLL distance between ground truth and predicted distribution, and the average cover rate (CR) of the estimated uncertainty interval in settings of 1) \texttt{SPCI}+base models, 2) \textbf{\name}+base models.}
\label{tab:uncertainty_coverage_rate}
\vspace{-0.1in}
\end{table}

\begin{table}[]
\small
\centering
\renewcommand{\arraystretch}{1.2}
\resizebox{0.6\linewidth}{!}{\begin{tabular}{|l||cc|}
\hline
Scheme  &  $\texttt{Without UC}$  & $\texttt{With UC}$    \\ \hline
\texttt{HiVT}+\textbf{\name} & 0.603 & 0.832  \\ \hline
\texttt{LaneGCN}+\textbf{\name} & 0.497 & 0.746 \\ \hline 
\texttt{LBA}+\textbf{\name} & 0.570 & 0.721 \\ \hline 
\texttt{LBF}+\textbf{\name} & 0.569 & 0.716 \\ \hline 
\end{tabular}}
\caption{The coverage rates of the uncertainty interval with and without the uncertainty calibration (UC) module.}
\vspace{-0.1in}
\label{tab:ablation_study_covrage_rate}
\end{table}

\begin{table}[]
\small
\centering
\renewcommand{\arraystretch}{1.2}
\vspace{-0.1in}
\resizebox{\linewidth}{!}{\begin{tabular}{|l||ccc|ccc|}
\hline
Scheme  &  ${\texttt{minADE}}_6 \downarrow$  & ${\texttt{minFDE}}_6 \downarrow$  & ${\texttt{MR}}_6 \downarrow$  & ${\texttt{minADE}}_1 \downarrow$  & ${\texttt{minFDE}}_1 \downarrow$  & ${\texttt{MR}}_1 \downarrow$   \\ \hline
\texttt{HiVT+\textbf{\name}} & 0.682 & 1.034 & 0.097 & 1.218 & 2.696 & 0.455  \\ \hline
\texttt{HiVT+DM} & 0.694 & 1.047 & 0.107 & 1.302 & 2.921 & 0.501 \\ \hline 
\end{tabular}}
\caption{Results of our \textbf{\name{}} and replacing the Gaussian process regression module with the self-attention based design for uncertainty estimation.}
\label{tab:ablation_study_GPR_VS_DM}
\end{table}

\subsection{Main Results} \label{subsec:main_results}
We adopt \texttt{HiVT}~\cite{liang2020learning}, \texttt{LaneGCN}~\cite{liang2020learning}, \texttt{LBA}~\cite{zhong2022aware}, \texttt{LBF}~\cite{zhong2022aware}, and \texttt{Trajectron++}~\cite{salzmann2020trajectron++} as base models and apply our \textbf{\name{}} on them to verify the efficacy of \textbf{\name{}}. To compare our \textbf{\name{}} with state-of-the-art methods, we also apply \texttt{SPCI}~\cite{xu2023sequential} into the above five base models to serve as the alternative of our \textbf{\name{}}. In specifically, we implement additional module whose structure is the same as the estimator in \texttt{SPCI} to predict the variance of the predicted trajectory and add the KLD~\cite{meyer2020learning} loss term to reduce the estimated uncertainty. As presented in Table.~\ref{tab:argoverse_v1^testing_results} and Table.~\ref{tab:uncertainty_coverage_rate}, our \textbf{\name{}} improves the prediction accuracy 7.07\% on average. Compared with incorporating \texttt{SPCI} in the base models, our \textbf{\name{}} reduces the output uncertainty which quantified by NLL by 25.41\% on average, and improves the average coverage rate of the estimated uncertainty interval by 21.02\%. 
The results indicate that our \textbf{\name{}} is capable of providing the trust worthy uncertainty quantification for the output trajectory of base model under potential distribution shift. Our estimated uncertainty intervals are narrow and achieve high coverage rate of covering the true target trajectories. 

Compared with the transformer-based model \texttt{HiVT}, applying our \textbf{\name{}} in \texttt{HiVT} provides the output uncertainty of \texttt{HiVT} by considering the distribution difference between the training and testing data. Compared with the base models of \texttt{LaneGCN}, \texttt{LBA}, \texttt{LBF}, and \texttt{Trajectron++} who provide a confidence score or the estimated distribution for each output trajectory, our \textbf{\name{}} provides the output distribution of base model instead of point estimates and calibrates the correspond uncertainty by taking into account the model performance on recent inputs. \texttt{SPCI} try to predict the conformal quantile by the learning-based estimator to provide the prediction interval. However, such estimator is prone to overfit the training data and invalid under distribution shift.

\subsection{Ablation Study} \label{subsec:ablation_study}
$\bullet$~\textbf{Conformal Prediction VS conformal P control module:} We conduct ablation study on replacing the P control uncertainty calibration module with the standard split conformal prediction~\cite{shafer2008tutorial}. More specificity, the standard split conformal prediction (CP) estimates the conformal quantile $\hat{q}$ as the $\frac{(1-\alpha)(N_2 + 1)}{N_2}$ smallest element in the conformal score set $\mathcal{S}={\{s\left(x_i, y_i\right)\}}_{i=1}^{N_2}$ from validation dataset. This conformal quantile $\hat{q}$ is only calculated once and fixed in the testing stage to calibrate the uncertainty.

We implement the standard split conformal prediction setting in base model \texttt{LaneGCN}. The average coverage rates of the uncertainty intervals estimated by our P control module and the split conformal prediction are 0.746 and 0.402, respectively. Our P control module outperforms the standard split conformal prediction. This helps to prove the effectiveness of our uncertainty calibration module in adapting to potential distribution shift. 

$\bullet$~\textbf{With VS without the P control uncertainty calibration module:} We further verify the efficacy of the P control uncertainty calibration module by removing the whole calibration process in the testing stage. As shown in Table.~\ref{tab:ablation_study_covrage_rate}, by calibrating the output uncertainty of the base model by the P control uncertainty calibration module, the coverage rate of the calibrated uncertainty interval improves 35.10\% on average comparing with without calibration. 

$\bullet$~\textbf{Gaussian process regression module VS direct modeling:} To validate the effectiveness of our Gaussian process regression module, we replace the Gaussian process regression module with the self-attention based design~\cite{mao2023leapfrog} to predict the variance of the output distribution and add the KLD loss term to reduce the estimated uncertainty during the training stage. We apply this setting on the base model \texttt{HiVT}. As shown in Table.~\ref{tab:ablation_study_GPR_VS_DM}, both of the methods achieve better prediction accuracy than without considering uncertainty in the base model. Our \textbf{\name} slightly exceed the direct modeling method in all prediction accuracy evaluation metrics. The results prove that our \textbf{\name} is sufficient in providing output uncertainty estimation for the trajectory prediction base model and improving the prediction accuracy. 



\section{Conclusion \& Discussion}\label{sec:conclusion_discussion}
In this study, we present the framework \textbf{\name} to estimate the output distribution for any existing trajectory prediction models under distribution shift, while improving the prediction accuracy and reducing the output uncertainty. \textbf{\name} adopts the Gaussian process regression module to model the output distribution of the base model. In addition, \textbf{\name} utilizes a statistical-based P control module to calibrate the estimated uncertainty by considering the performance limitation of base model upon recent inputs. The experiment results demonstrate the efficacy of \textbf{\name} in improving the prediction accuracy and reducing the prediction uncertainty. Our findings highlight the importance of quantifying the uncertainty of output trajectory in trajectory prediction under distribution shift. In the future work, we plan to extend our work to more state-of-the-art models, provide uncertainty of output trajectory in each time step, and further compare the performance difference of estimating uncertainty and output distribution by different designs.

\section{Acknowledgments}
The work is supported by the National Science Foundation under Grants CNS-2047354 (CAREER), and the
New England University Transportation Center (NEUTC). Funding for the NEUTC Program is provided by the Office of Assistant Secretary for Research and Innovation (OST-R) of the United States Department of Transportation. The recommendations of this study are those of the authors and do not represent the views of NEUTC.

\bibliography{aaai25}

\end{document}